\documentclass[10pt,twocolumn,letterpaper]{article}

\usepackage[pagenumbers]{cvpr} 

\usepackage{graphicx}
\usepackage{amsmath}
\usepackage{amssymb}
\usepackage{booktabs}
\usepackage{comment}
\usepackage{caption}

\usepackage[pagebackref,breaklinks,colorlinks]{hyperref}

\usepackage[capitalize]{cleveref}
\crefname{section}{Sec.}{Secs.}
\Crefname{section}{Section}{Sections}
\Crefname{table}{Table}{Tables}
\crefname{table}{Tab.}{Tabs.}

\newcommand{\rnum}[1]{\lowercase\expandafter{\romannumeral #1\relax}}

\def\cvprPaperID{8400} 
\def\confName{CVPR}
\def\confYear{2023}

\begin{document}

\title{ProphNet: Efficient Agent-Centric Motion Forecasting with \\Anchor-Informed Proposals} 

\author{
  Xishun Wang \quad Tong Su \quad Fang Da \quad Xiaodong Yang\thanks{Corresponding author \texttt{xiaodong@qcraft.ai}}\\
  QCraft\\
}

\maketitle

\begin{abstract}
Motion forecasting is a key module in an autonomous driving system. Due to the heterogeneous nature of multi-sourced input, multimodality in agent behavior, and low latency required by onboard deployment, this task is notoriously challenging. To cope with these difficulties, this paper proposes a novel agent-centric model with anchor-informed proposals for efficient multimodal motion prediction. We design a modality-agnostic strategy to concisely encode the complex input in a unified manner. We generate diverse proposals, fused with anchors bearing goal-oriented scene context, to induce multimodal prediction that covers a wide range of future trajectories. Our network architecture is highly uniform and succinct, leading to an efficient model amenable for real-world driving deployment. Experiments reveal that our agent-centric network compares favorably with the state-of-the-art methods in prediction accuracy, while achieving scene-centric level inference latency.

\end{abstract}

\section{Introduction}
\label{sec:intro}
Predicting the future behaviors of various road participants is an essential task for autonomous driving systems to be able to safely and comfortably operate in dynamic driving environments. However, motion forecasting is extremely challenging in the sense that (\rnum{1}) the input consists of interrelated modalities from multiple sources; (\rnum{2}) the output is inherently stochastic and multimodal; and (\rnum{3}) the whole prediction pipeline must fulfill tight run time requirements with a limited computation budget.

A motion forecasting model typically collects the comprehensive information from perception signals and high-definition (HD) maps, such as traffic light states, motion history of agents, and the road graph~\cite{simtrack,pillar-motion,light,hdmap}. Such a collection of information is a heterogeneous mix of static and dynamic, as well as discrete and continuous elements. Moreover, there exist complex semantic relations between these components, including agent-to-agent, agent-to-road, and road-to-road interactions. Previous methods in the field usually model this diverse set of input via an equally intricate system with modality-specific designs. LaneGCN~\cite{liang2020learning} adopts four sub-networks to separately process the interactions of agent-to-agent, agent-to-road, road-to-agent and road-to-road. MultiPath++ is developed in~\cite{mp++} to employ multi-context gating to first capture the interactions between a target agent and other agents, and then fuse them with the map in a hierarchical manner. 

\begin{figure}[t]
        \centering
        \includegraphics[width=\linewidth]{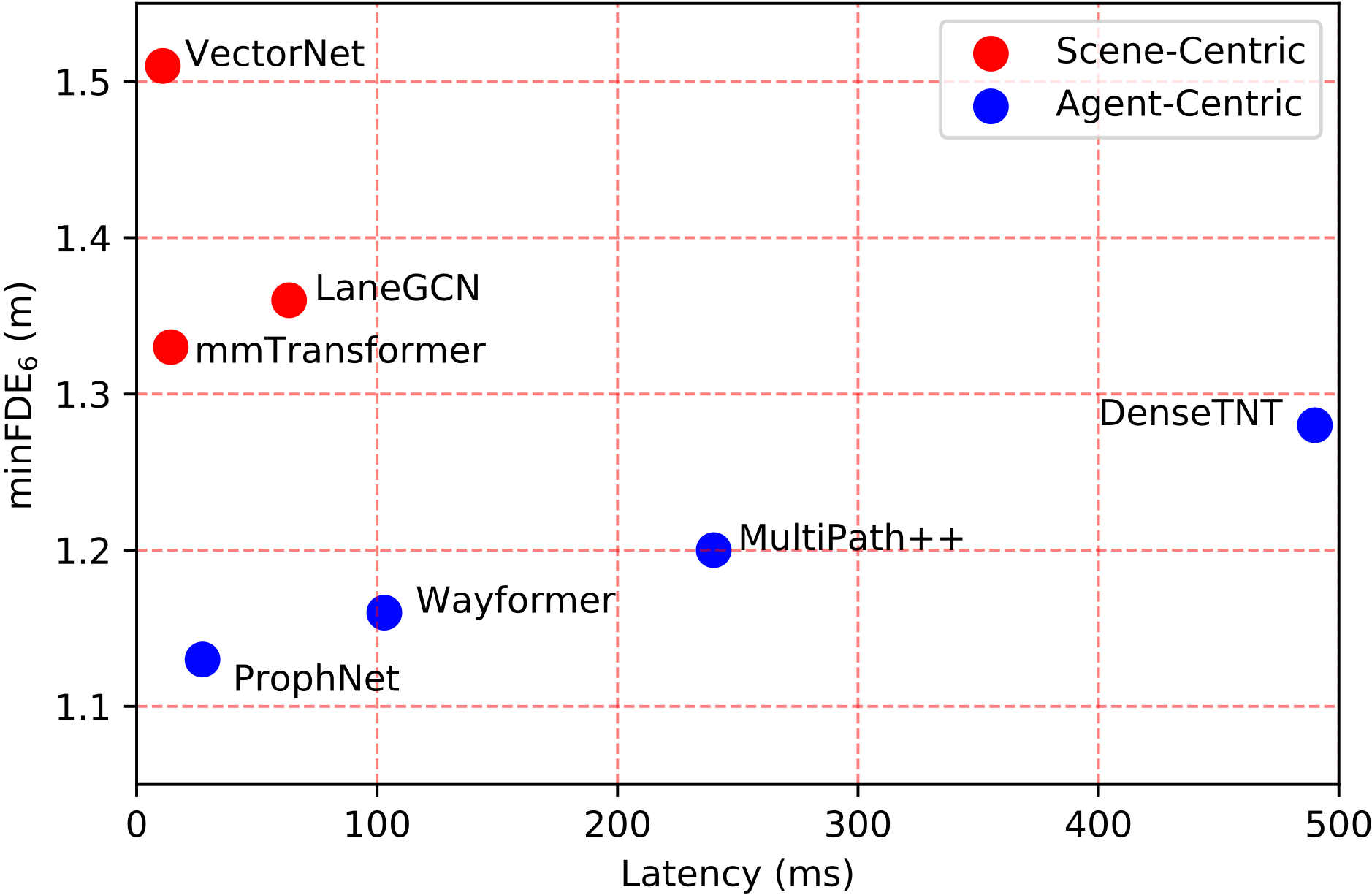}
        \caption{Overview of accuracy-and–latency trade-off for the task of motion forecasting on Argoverse-1. ProphNet outperforms the state-of-the-art methods in prediction accuracy and considerably speeds up the agent-centric inference latency, leading to the best balance between accuracy and latency.} 
        \label{intro}
\end{figure}

Due to the unknown intent of an agent, motion prediction output is highly multimodal by nature. It is often impossible to assert with full certainty whether a vehicle will go straight or turn right as it approaches an intersection. The motion prediction model is therefore required to accurately model the underlying probability distribution of future behaviors. 
Recently, there have been various efforts on improving output multimodality. 
TNT~\cite{tnt} introduces three stages including goal (endpoint of a predicted trajectory) prediction, goal-conditioned motion estimation, and trajectory scoring. 
DenseTNT is proposed in~\cite{densetnt} to improve upon the former by predicting from dense goal candidates and conducting an iterative offline optimization algorithm to generate multi-future pseudo labels. 
A region based training strategy is developed in mmTransformer~\cite{liu2021multimodal} to manually partition a scene into subregions to enforce predictions to fall into different regions to promote multimodality.

In a practical autonomous driving system, the whole prediction pipeline typically operates at a frequency of 10Hz, including (\rnum{1}) feature extraction from upstream modules, (\rnum{2}) network inference, and (\rnum{3}) post-processing. Therefore, a motion forecasting network is required to meet the strict inference latency constraint to be useful in the real-world setting. 
While the recent agent-centric paradigm~\cite{mp++, wayformer} is trending and has made remarkable progress on improving accuracy, its latency is dramatically increased compared to its scene-centric counterpart~\cite{vectornet, liu2021multimodal}, as shown in Figure~\ref{intro}. 
This is because agent-centric models compute scene representations with respect to each agent separately, meaning the amount of features to process is dozens of times larger than that with scene-centric models. 
This substantially increases the latency of the whole prediction pipeline and is computationally prohibitive for onboard deployment.

In light of these observations, we present \textbf{ProphNet}: an efficient agent-centric motion forecasting model that fuses heterogeneous input in a unified manner and enhances multimodal output via a design of anchor-informed proposals. In contrast to the existing complex modality-specific processing~\cite{liang2020learning, mp++}, we develop a modality-agnostic architecture that combines the features of agent history, agent relation and road graph as the agent-centric scene representation (AcSR), which directly feeds to a unified self-attention encoder~\cite{transformer}. In this way, we motivate the self-attention network to learn the complex interactions with minimum inductive bias. Additionally, we propose the anchor-informed proposals (AiP) in an end-to-end learning fashion to induce output multimodality. In our approach, \textit{proposals} are the future trajectory embeddings conjectured exclusively from agent history, and \textit{anchors} refer to the goal embeddings learned from AcSR. In such a manner, the network learns to first generate proposals to maximize diversity without environmental restrictions, and then select and refine after absorbing anchors that carry rich goal-oriented contextual information. 
Based on random combinations of AiP, we further introduce the hydra prediction heads to encourage the network to learn complementary information and meanwhile perform ensembling readily.

\begin{figure*}[t]
        \centering
        \includegraphics[width=\textwidth]{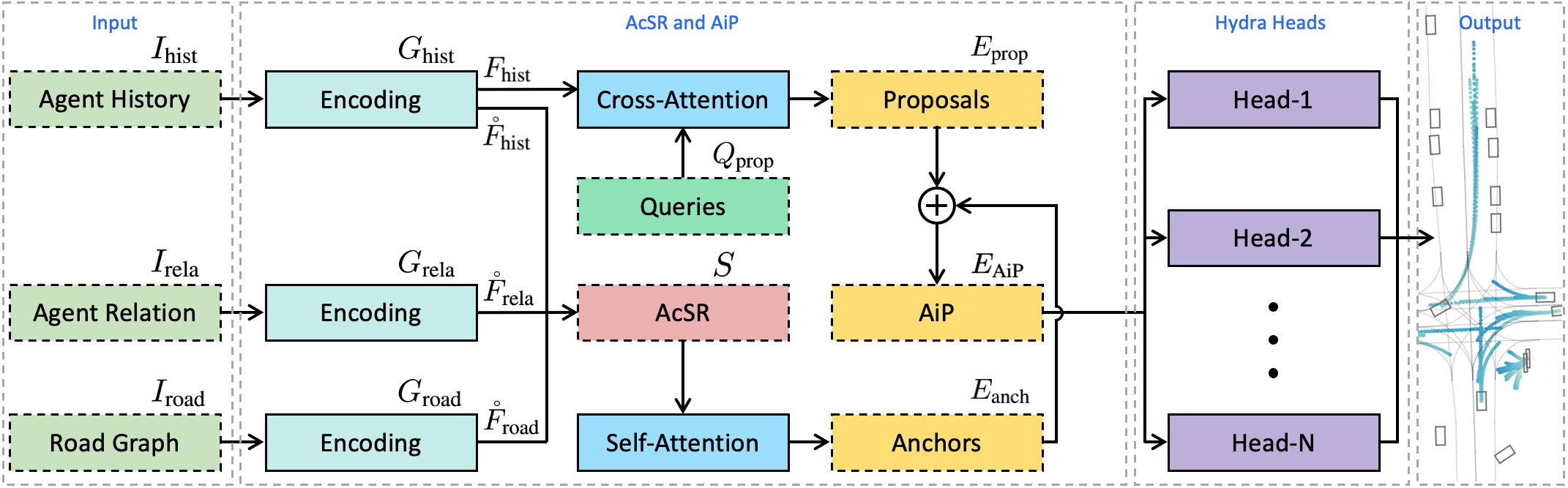}
        \caption{A schematic overview of ProphNet. We uniformly encode the heterogeneous input by gMLP, and combine the compact encoding features to form a unified representation space AcSR. We generate proposals through cross-attention between learnable queries and full history encoding. Anchors are learned based on self-attention of AcSR. We introduce AiP by integrating proposals and anchors, and randomly select subsets from AiP to feed into hydra prediction heads to output the final prediction. We use the solid and dashed boxes to indicate operators and operands in the network, respectively.}
        \label{arch}
\end{figure*}

As illustrated in Figure~\ref{arch}, this design formulates a succinct network with high efficiency in terms of both architecture and inference. Instead of employing a self-attention encoder for each input modality~\cite{liu2021multimodal, liang2020learning}, the self-attention on AcSR unifies the learning of intra- and inter-modality interactions in a single compact space, which allows the network to assign associations within and across input modalities with maximum flexibility. Our model also considerably reduces inference latency and performs prediction with a peak latency as low as 28ms. In addition, rather than heuristically predefining or selecting goals to strengthen output multimodality~\cite{tnt, densetnt, mp}, AiP is learned end-to-end to infuse goal based anchors to proposals that are deliberately produced with diverse multimodality. In the end, together with the hydra prediction heads, our network is capable of generating future trajectories with rich variety.

Our main contributions are summarized as follows. First, we develop an input-source-agnostic strategy based on AcSR to model heterogeneous input and simplify network architecture, making the model amenable for real-world driving deployment. Second, we propose a novel framework that couples proposal and anchor learning end-to-end through AiP to promote output multimodality. Third, we introduce hydra prediction heads to learn complementary prediction and explore ensembling. Fourth, we show that our network, as an agent-centric model, achieves state-of-the-art accuracy on the popular benchmarks while maintaining scene-centric level low inference latency.

\section{Related Work}

\noindent\textbf{Representation Paradigms.} Motion forecasting methods can be roughly categorized into two types according to the spatial feature representation: rasterized~\cite{mp, covernet, mfp} and vectorized~\cite{vectornet, densetnt, dcms}. Rasterized methods render agent states and HD maps as color-coded attributes, and use image-oriented networks (e.g., CNN) to encode scene context. Vectorized methods represent agent dynamics and map elements in their vectorized form. Combined with the attention mechanism, vectorized approaches have dominated the leading results recently.
On the other hand, regarding the reference points for feature encoding, methods fall into two groups: scene-centric~\cite{scenetransformer, vectornet} and agent-centric~\cite{tnt, wayformer}. In the former, scene representation is extracted once for all agents in a shared coordinate frame, while the latter repeatedly builds a normalized coordinate frame for each agent and extracts corresponding scene features. As a result, the scene-centric models are generally more computationally efficient, and the agent-centric models deliver better accuracy.

\noindent\textbf{Modeling Heterogeneous Input.} Modeling the complex interactions among diverse types of input is vital for effectively capturing the scene context. LaneGCN~\cite{liang2020learning} uses graph convolution to improve road graph representation, and proposes four input-specific fusion modules for interaction modeling. Based on LaneGCN, BANet~\cite{banet} further incorporates more road graph elements and interaction fusion blocks, while PAGA~\cite{paga} augments map encoding with attention between non-adjacent map elements. SceneTransformer~\cite{scenetransformer} develops multi-axis attention to model spatial and temporal information, where agent-to-agent and agent-to-road interactions are learned in separate modules. Wayformer~\cite{wayformer} also adopts multi-axis attention and studies fusing multi-sourced input at different stages. Most previous works focus on explicitly processing and combining individual input sources to better model their interactions, thus inevitably resulting in increased model complexity.

\noindent\textbf{Enhancing Multimodal Output.} Extensive research efforts have been made in multimodal prediction. Planning based methods~\cite{pip} apply basic planning rules to generate plenty of feasible trajectories following the road graph and kinematic constraints as trajectory candidates. In~\cite{liu2021multimodal} mmTransformer uses a region based training scheme placing predictions in manually defined regions to enrich the output. MultiPath~\cite{mp} clusters trajectories from training data as prediction anchors to prevent mode collapse. MultiPath++~\cite{mp++} learns a fixed set of hidden anchor embeddings to capture different modalities. TNT~\cite{tnt} and DenseTNT~\cite{densetnt} rely on sampling sparse and dense goal candidates to achieve good coverage. We aim to induce multimodality through the proposed anchor-informed proposals that can be learned fully end-to-end and is capable of adapting to specific input.

\section{Method}

Figure~\ref{arch} illustrates the overall architecture of ProphNet, which involves encoding heterogeneous input into an agent-centric scene representation, end-to-end generation of proposals and anchors, coupling them to form anchor-informed proposals, as well as hydra prediction heads to produce multimodal and complementary trajectories.

\subsection{Input Representation}

ProphNet performs motion forecasting based on multi-sourced input, including the motion history of target agent, other agents in the surroundings, and road graph. Representations of different input are detailed as follows.

\noindent\textbf{Agent History.} It is a state sequence consisting of position, velocity and heading of the target agent at each observation step, i.e., $I_{\text{hist}} = [(x_t, y_t, v^x_{t}, v^y_{t}, \sin\theta_t, \cos\theta_t)_{t=-T+1:0}]$. An agent-centric coordinate frame is defined by aligning the origin with the last observed position of a target agent and the x+ axis with its last observed heading, such that the target agent is located at the origin and heading to the east at the current time $t=0$. The historical states of the target agent are then transformed to this coordinate frame. We repeat this coordinate frame normalization for all target agents to be predicted in a scene.

\noindent\textbf{Agent Relation.} This describes the relative states of other agents with respect to a target agent. 
Again, the historical states of other agents are transformed to the agent-centric coordinate frame of the target agent. We take into account the neighboring agents within a certain range (related to prediction horizon). 
Assuming \(B\) agents are considered, the agent relation is \(I_{\text{rela}}=\{(I^i_{\text{rela}})_{i=1:B}\}\), where \(I^i_{\text{rela}}\), similar to $I_{\text{hist}}$, is a sequence of position, velocity and heading of the $i$-th neighboring agent relative to the target agent. 

\noindent\textbf{Road Graph.} This defines the map area where the target agent may reach in the prediction horizon. Similarly, the map elements are transformed to the agent-centric coordinate frame. We exploit lane centerlines, lane boundaries and traffic controls (e.g., turns and stop signs) to represent the road graph, and approximate each lane centerline or lane boundary by a polyline, which is an ordered set of connected line segments with equal length.
A road graph with $L$ polylines is represented by $I_{\text{road}} = \{(I^i_{\text{road}})_{i=1:L}\}$, where $I^i_{\text{road}}$ is the $i$-th polyline.
Let $(a, b)$ be the endpoints of a segment, $\alpha$ be the heading angle, and $c$ be the closest point on the segment to the origin. We represent a line segment by $[(a-b) / \|a-b\|_2, \|c\|_2, c/ \|c\|_2, \|a-c\|_2, \alpha]$.

\subsection{Agent-Centric Scene Representation (AcSR)}
We refer to the collection of representation encoding of the heterogeneous input as AcSR. It provides a unified foundation for learning interactions across all input sources. 
We employ multilayer perceptron with gating (gMLP)~\cite{gmlp}, a simple yet effective method to model sequential data, to perform uniform encoding on various input sources.

\noindent\textbf{Encoding Agent History.} We adopt a gMLP $G_{\text{hist}}$ to encode the historical states of a target agent as $F_{\text{hist}} = [f_{-T+1}, ..., f_{0}] = G_{\text{hist}}(I_{\text{hist}})$, where $F_{\text{hist}} \in \mathbb{R}^{T \times D_F}$ is a temporally ordered sequence of encoded features. We define $F_{\text{hist}}$ as the full history encoding, and $\mathring{F}_{\text{hist}} = f_0$ as the compact history encoding that only contains the feature at the current time step $t=0$. $F_{\text{hist}}$ can be adopted as a memory to generate diverse proposals as it captures the complete spatio-temporal cues, while $\mathring{F}_{\text{hist}}$ can be thought of as a compact representation to constitute the set of AcSR. 

\noindent\textbf{Encoding Agent Relation.} We use another gMLP $G_{\text{rela}}$ to encode the relative states of neighboring agents with respect to a target agent as $F_{\text{rela}} = \{(F^i_{\text{rela}})_{i=1:B}\} = G_{\text{rela}}(I_{\text{rela}})$, where $F^i_{\text{rela}} \in R^{T \times D_F}$ is a temporally ordered sequence of encoded features of the $i$-th neighboring agent. All neighboring agents of a target agent share the same gMLP $G_{\text{rela}}$. Similarly, $\mathring{F}_{\text{rela}} = \{(\mathring{F}^i_{\text{rela}})_{i=1:B}\}$ is used to form the compact agent relation encoding for AcSR.

\noindent\textbf{Encoding Road Graph.} We utilize a third gMLP $G_{\text{road}}$ to encode the polylines in the road graph around a target agent as $F_{\text{road}} = \{(F^i_{\text{road}})_{i=1:L}\} = G_{\text{road}}(I_{\text{road}})$, where $F^i_{\text{road}} \in \mathbb{R}^{T \times D_F}$ is an ordered sequence of encoded segments of the $i$-th polyline. All polylines in the road graph share the same gMLP $G_{\text{road}}$. We also collect the compact polyline encoding of a road graph to compose $\mathring{F}_{\text{road}} = \{(\mathring{F}_{\text{road}}^i)_{i=1:L}\}$ as the local map representation in AcSR.

We combine the compact encoded features from the three aforementioned input sources to construct the AcSR, i.e., $S = \{\mathring{F}_{\text{hist}}, \mathring{F}_{\text{rela}}, \mathring{F}_{\text{road}}\}$. AcSR provides a single unified feature space for the complex interaction learning, leading to a succinct network architecture. Furthermore, the different input representations are uniformly encoded by gMLP, which is beneficial for reducing the gap between heterogeneous information and enables the following attention operations to better learn the interactions within the same source and across different sources alike.

\subsection{Proposal Generation}
\label{proposal-generation}

To enhance the output multimodality, our network is designed to first learn to generate proposals merely from the motion history of a target agent. This maximizes the diversity of future trajectory candidates by waiving the constraints from the particular driving scenarios. 
We set the number of generated proposals to $K$, which is sufficiently larger than the required number of output modality $M$. A set of $K$ learnable proposal queries $Q_{\text{prop}}$ is applied to attend to the full history encoding $F_{\text{hist}}$ through cross-attention to produce the set of proposals $E_{\text{prop}} = \{(E^i_{\text{prop}})_{i=1:K}\} \in \mathbb{R}^{K \times D_E}$, as demonstrated in Figure~\ref{arch}. It can be verified that, for the proposal generation, using $\mathring{F}_{\text{hist}}$ results in inferior performance than $F_{\text{hist}}$, which retains the full historical spatial and temporal information of the target agent (Section~\ref{abl}). To further encourage the diversity of proposals, $Q_{\text{prop}}$ vectors are initialized orthogonally.

\subsection{Anchor Learning}

Anchors are learned end-to-end in the network to convey goal-oriented environmental information, while preserving diversity. 
We set the number of anchors to $K$, same as the number of proposals, so that each anchor corresponds to one proposal.
We utilize self-attention on AcSR to let the heterogeneous elements in $S$ attend to each other and learn related interactions with maximum flexibility. 
Assuming the output of self-attention is $\{(R_i)_{i=1:1+B+L}\}$, the target agent representation $R_1$ has fused the input context.
Our network is then supervised to learn to predict $K$ trajectory endpoints based on $R_1$.
And the predicted endpoint that is closest to the ground truth is chosen as the correct one. 
A smooth $\ell_1$ loss is optimized to minimize the distance between ground truth and correct prediction.

Our model does not directly adopt the predicted endpoints, but make use of their embeddings (i.e., the pre-output features) as the anchors $E_{\text{anch}} = \{(E^i_{\text{anch}})_{i=1:K}\} \in \mathbb{R}^{K \times D_E}$, which are utilized to inject the goal-oriented scene context to the generated proposals. This is a new genre of anchors compared to those used in previous works.    
In TNT~\cite{tnt}, anchors are uniformly sampled from the map using hand-crafted rules. 
In MultiPath~\cite{mp}, anchors are a set of pre-defined trajectories clustered from training data. 
In MultiPath++~\cite{mp++}, anchors are learnable model parameters that are fixed after training and independent of input. 
As a contrast, we propose to exploit the anchor embeddings to facilitate trajectory learning.
Compared to TNT and MultiPath, our anchors are obtained more adaptively and conveniently via end-to-end learning. Compared to MultiPath++, our anchors correspond to individual samples, thus carrying concrete sample-specific information.

\subsection{Anchor-Informed Proposals (AiP)}

As mentioned above, proposals can be viewed as unconstrained future trajectories inferred solely from agent history, and anchors convey goal based contextual information. Here we introduce AiP to enable the network for further selection and refinement, while achieving diverse output multimodality. 
One way to fuse proposals and anchors is through attention to use proposals to query anchors. However, we find this results in a collapse of multimodality to some extent. Instead, we directly sum proposals and anchors together to generate AiP, which is found to be simple yet effective to accomplish the fusion in our experiments:         
\begin{equation}
E_{\text{AiP}} = E_{\text{prop}} + E_{\text{anch}}, 
\label{inf}
\end{equation}
where $E_{\text{AiP}} \in \mathbb{R}^{K \times D_E}$ is provided to the following hydra prediction heads to output the final predicted trajectories.

\subsection{Hydra Prediction Heads}

As discussed in Section~\ref{proposal-generation}, we intentionally generate more proposals than the required number of output modality (i.e., $K > M$). In practice, we set $K = 2M$, and randomly select $M$ embeddings from $E_{\text{AiP}}$ as a subset. This random selection is repeated $N$ times, and thus $N$ different subsets can be obtained. Accordingly, we construct $N$ heads as the hydra prediction heads, each of which takes as input one subset and produces multimodal future trajectories. With the hydra heads and associated subsets that are randomly selected, the output modality can be further enhanced. In addition, this design also paves the way for ensembling by encouraging the hydra heads to learn complementary prediction. Our single network obtains a group of $N \times M$ trajectories via ensembling the predicted results from the hydra prediction heads. We then simply employ non-maximum suppression to merge them into the final prediction output with $M$ multimodal trajectories.

We parameterize the distribution of each predicted trajectory as a Gaussian mixture model (GMM): 
\begin{equation}
p(\tau)=\sum_{i=1}^{M} p^i \prod_{t=1}^H \mathcal{N}(\tau_t-\mu_t^i, \Sigma_t^i), 
\label{distr}
\end{equation}
where $\tau$ is a trajectory in the prediction horizon $H$, $p^i$ is the likelihood of the $i$-th mode, $\tau_t \in \mathbb{R}^2$ is the trajectory point at time step $t$ with predicted mean $\mu^i_t \in \mathbb{R}^2$ and covariance $\Sigma^i_t \in \mathbb{R}^{2 \times 2}$.
For network training, we maximize the likelihood of ground truth under the predicted trajectory distribution. We apply hard labeling to assign the trajectory with the smallest endpoint distance to ground truth as positive. And only the errors between positive trajectories and ground truth are considered in the regression loss.

\section{Experiments}

In this section, we first describe our experimental setup including datasets, evaluation metrics and implementation details. A variety of ablation studies are conducted to understand the contribution of each individual design in our network. We report extensive comparisons with the state-of-the-art methods on two popular benchmarks.

\subsection{Datasets and Metrics}

We evaluate our approach on two large-scale motion forecasting datasets: Argoverse-1~\cite{argo1} and Argoverse-2~\cite{argo2}. \textbf{Argoverse-1} contains 333K real-world driving sequences selected from intersections or dense traffic. 
Each sequence consists of 2 seconds of history and 3 seconds of future, sampled at 10Hz. Following the official split, the training, validation and test sets have 205,942, 39,472 and 78,143 sequences respectively. \textbf{Argoverse-2} upgrades the first version on scenario complexity, prediction horizon and agent types. It contains 250K sequences mined for challenging scenes with rich geographic diversity. This new dataset extends the prediction horizon to 6 seconds, and covers 5 types of agents including vehicles, pedestrians, cyclists, motorcyclists and buses. There are 200,000, 25,000 and 25,000 sequences in training, validation and test sets.

We adopt the official evaluation metrics defined by the two benchmarks: minADE, minFDE, brier-minFDE, and MR. minADE (average displacement error) is the minimum of the time-average distance between predicted trajectories and the ground truth trajectory, over $M$ prediction modes. minFDE (final displacement error) is the minimum distance between $M$ predicted endpoints and the ground truth endpoint. brier-minFDE additionally takes the prediction probability into account. MR (miss rate) is the fraction of scenes where none of $M$ predicted endpoints are within a certain distance to ground truth.

\begin{table}[t]
  \centering
  \resizebox{\columnwidth}{!} 
  {
  \begin{tabular}{@{}lc@{}lc@{}lc@{}lc@{}lc@{}lc@{}lc@{}}
    \toprule
    Method & minADE$_6$& &minFDE$_6$& &minADE$_1$& &minFDE$_1$ \\ 
    \midrule
    VectorNet~\cite{vectornet}  &-& &-& &1.66& &3.67\\
    LaneRCNN~\cite{lanercnn}  &0.77& &1.19& &1.33& &2.85\\
    TPCN~\cite{tpcn}   &0.73& &1.15& &1.34& &2.95\\
    mmTransformer~\cite{liu2021multimodal}   &0.71& &1.15& &-& &-\\
    LaneGCN~\cite{liang2020learning}    & 0.71 & & 1.08& &1.35& &2.97\\
    PAGA~\cite{paga} & 0.69 & & 1.02& &1.31& &2.87\\
    \midrule
    MultiPath~\cite{mp}   & 0.80 & & 1.68& &-& &- \\
    TNT~\cite{tnt}  & 0.73 & & 1.29& &-& &-\\
    DenseTNT~\cite{densetnt}  & 0.73 & & 1.05& &-& &-\\
    ProphNet-S  & \textbf{0.68} & & \textbf{0.97}& &\textbf{1.28}& &\textbf{2.77}\\
    \bottomrule
  \end{tabular}
  }
  \caption{Comparison of ProphNet and the state-of-the-art methods on the validation set of Argoverse-1. Groups 1 and 2 are the scene-centric and agent-centric methods.}
  \label{tab:Argo1-val}
\end{table}

\subsection{Implementation Details}

We implement ProphNet in PyTorch~\cite{NEURIPS2019_9015} and train on 16 NVIDIA V100 GPUs with a batch size of 64. 
All gMLPs including $G_{\text{hist}}$, $G_{\text{rela}}$ and $G_{\text{road}}$ have the same hidden layer dimension 256. 
For all attention layers, the multi-head attention is configured to 4 heads, and the dimension of each head is 64. We employ one cross-attention layer for generating 
$E_{\text{prop}}$ and three self-attention layers for $E_{\text{anch}}$. 
In all experiments, the numbers of output modality, AiP and hydra prediction heads are set as $M = 6$, $K = 12$ and $N =3$, respectively. 
We use Adam~\cite{adam} to train the model for 60 epochs, and set the initial learning rate to $2 \times 10^{-4}$ and decay the learning rate to $2 \times 10^{-5}$ after 50 epochs. 
We keep the same setting for both datasets.

\begin{table*}[t]
  \centering
  \begin{tabular}{@{}lc@{}lc@{}lc@{}lc@{}lc@{}lc@{}lc@{}lc@{}lc@{}lc@{}lc@{}}
    \toprule
    Method &minADE$_6$& &minFDE$_6$& &minADE$_1$& &minFDE$_1$& &MR& &brier-minFDE\\
    \midrule
    LaneRCNN~\cite{lanercnn}  &0.9038& &1.4526& &1.6852& &3.6916& &0.1232& &2.1470\\
    LaneGCN~\cite{liang2020learning}  &0.8703& &1.3622& &1.7019& &3.7624& &0.1620& & 2.0539 \\
    mmTransformer~\cite{liu2021multimodal}  &0.8436& &1.3383& &1.7737& &4.0033& &0.1540& &2.0328 \\
    TPCN~\cite{tpcn}  &0.8153& &1.2442& &1.5752& &3.4872& &0.1333& &1.9286\\
    SceneTransformer~\cite{scenetransformer}  &0.8026& &1.2321& &1.8108& &4.0551& &0.1255& &1.8868\\
    \midrule
    TNT~\cite{tnt} &0.9097& &1.4457& &2.1740& &4.9593& &0.1656& &2.1401\\
    DenseTNT~\cite{densetnt} &0.8817& &1.2815& &1.6791& &3.6321& &0.1258& &1.9759\\
    MultiPath++~\cite{mp++} &0.7897& &1.2144& &1.6235& &3.6141& &0.1324& &1.7932 \\
    Wayformer~\cite{wayformer} &\textbf{0.7676}& &1.1616& &1.6360& &3.6559& &0.1186& &1.7408 \\
    ProphNet &0.7726& &\textbf{1.1442}& &\textbf{1.5240}& &\textbf{3.3341}& &\textbf{0.1121}& &\textbf{1.7323} \\
    \bottomrule
  \end{tabular}
  \caption{Comparison of ProphNet (single model without ensembling) and the state-of-the-art methods on the test set of Argoverse-1. Groups 1 and 2 are the scene-centric and agent-centric methods.}
  \label{tab:Argo1ranking}
\end{table*}

\begin{table*}[t]
  \centering
  \begin{tabular}{@{}lc@{}lc@{}lc@{}lc@{}lc@{}lc@{}lc@{}lc@{}lc@{}lc@{}lc@{}}
    \toprule
    Method &minADE$_6$& &minFDE$_6$& &minADE$_1$& &minFDE$_1$& &MR& &brier-minFDE\\
    \midrule
    GOHOME~\cite{gohome}  &0.88& &1.51& &1.95& &4.71& &0.20& &2.16 \\
    GoRela~\cite{gorela}  &0.76& &1.48& &1.82& &4.62& &0.22& &2.01 \\
    TENET~\cite{tenet}  &0.70& &1.38& &1.84& &\textbf{4.69}& &0.19& &1.90 \\
    ProphNet &\textbf{0.68}& &\textbf{1.33}& &\textbf{1.80}& &4.74& &\textbf{0.18}& &\textbf{1.88} \\
    \bottomrule
  \end{tabular}
  \caption{Comparison of ProphNet (single model without ensembling) and the state-of-the-art methods on the test set of Argoverse-2.}
  \label{tab:Argo2}
\end{table*}

\subsection{Results on Argoverse-1}
We start from the comparison of ProphNet and the leading algorithms on the validation set of Argoverse-1. Here we do not use ensembling for a fair comparison against the methods without such a design. So we disable the hydra prediction heads and employ the single-head version, i.e., ProphNet-S. As demonstrated in Table~\ref{tab:Argo1-val}, ProphNet-S outperforms other methods by a large margin. We attribute our superior result of minFDE to the proposed AiP, which fuses diversified proposals with goal-oriented anchors. And the anchor learning is explicitly trained to estimate the feasible goals, which is reflected in the more accurate endpoint prediction. This comparison clearly shows the advantage of the overall design of our approach.

We then evaluate ProphNet on the test set of Argoverse-1. We use dropout augmentation on agent relation and road graph, where 10\% elements are randomly dropped. We compare with the results of published methods in Table~\ref{tab:Argo1ranking}. 
Compared to the leading scene-centric methods, ProphNet demonstrates strong advantage over all evaluation metrics. As for the comparison with most recent agent-centric methods, MultiPath++~\cite{mp++} employs 5 heads and each produces 64 trajectories, while ProphNet uses 3 heads and each with 6 output trajectories. Wayformer~\cite{wayformer} trains 15 models for ensembling. In contrast, our prediction result is generated by a single model. It is observed that ProphNet significantly outperforms MultiPath++, and overall performs better than Wayformer. In addition to improving the prediction accuracy, ProphNet is computationally efficient. See the details of latency comparison and analysis in Section~\ref{sec:latency}.

\subsection{Results on Argoverse-2}
We also evaluate ProphNet on the test set of the latest benchmark Argoverse-2. 
We use the same configuration as that of Argoverse-1. 
Compared to the results of published methods, ProphNet achieves superior performance, in particular over TENET~\cite{tenet}, which is the winner of the motion forecasting challenge on Argoverse-2 in 2022.

\subsection{Latency Analysis}
\label{sec:latency}
In addition to prediction accuracy, we take a closer look into inference latency, which is an equally important measurement for practical onboard deployment. We compare the inference latency of ProphNet with the state-of-the-art methods in Table~\ref{tab:latency}. 
We report the inference latency on a single NVIDIA V100 GPU and set the number of agents to 64, a reasonable amount of neighboring agents in practice. For the scene-centric models, we first compute the average latency per agent. 
In general, the scene-centric models run considerably faster than most agent-centric models, often by an order of magnitude. In the contrary, as an agent-centric model, ProphNet achieves the inference latency around 28ms, significantly speeding up run time to the scene-centric level. 
We attribute our efficient inference to the uniform and succint model deisgn, as well as ruling out the computationally heavy operations such as multi-axis~\cite{scenetransformer, wayformer} and autoregressive sequential prediction~\cite{scenetransformer, tenet}. It is also observed that the hydra prediction heads in our approach only increase the latency slightly, from 27.4ms to 28.0ms.

\begin{table}
  \centering
  \begin{tabular}{@{}lc@{}lc@{}lc@{}}
    \toprule
    Method & Latency (ms) &  &GFLOPs\\
    \midrule
    VectorNet~\cite{vectornet}  & 10.9 & &0.01\\
    LaneGCN~\cite{liang2020learning}  & 63.4 &  &0.13\\
    mmTransformer~\cite{liu2021multimodal}  & 14.2& &0.01\\
    \midrule
    DenseTNT~\cite{densetnt}  & 490.1 & &0.62\\
    MultiPath++~\cite{mp++}  & 240.4 &  &2.19\\
    Wayformer~\cite{wayformer}  & 102.3 & &2.13\\
    ProphNet-S & 27.4 & &0.39\\
    ProphNet & 28.0 & &0.40\\
    \bottomrule
  \end{tabular}
  \caption{Comparison of ProphNet and its variant to the state-of-the-art methods on inference latency and number of GFLOPs. Groups 1 and 2 are the scene-centric and agent-centric methods.}
  \label{tab:latency}
\end{table}

\begin{figure}[t]
        \centering
        \includegraphics[width=\linewidth]{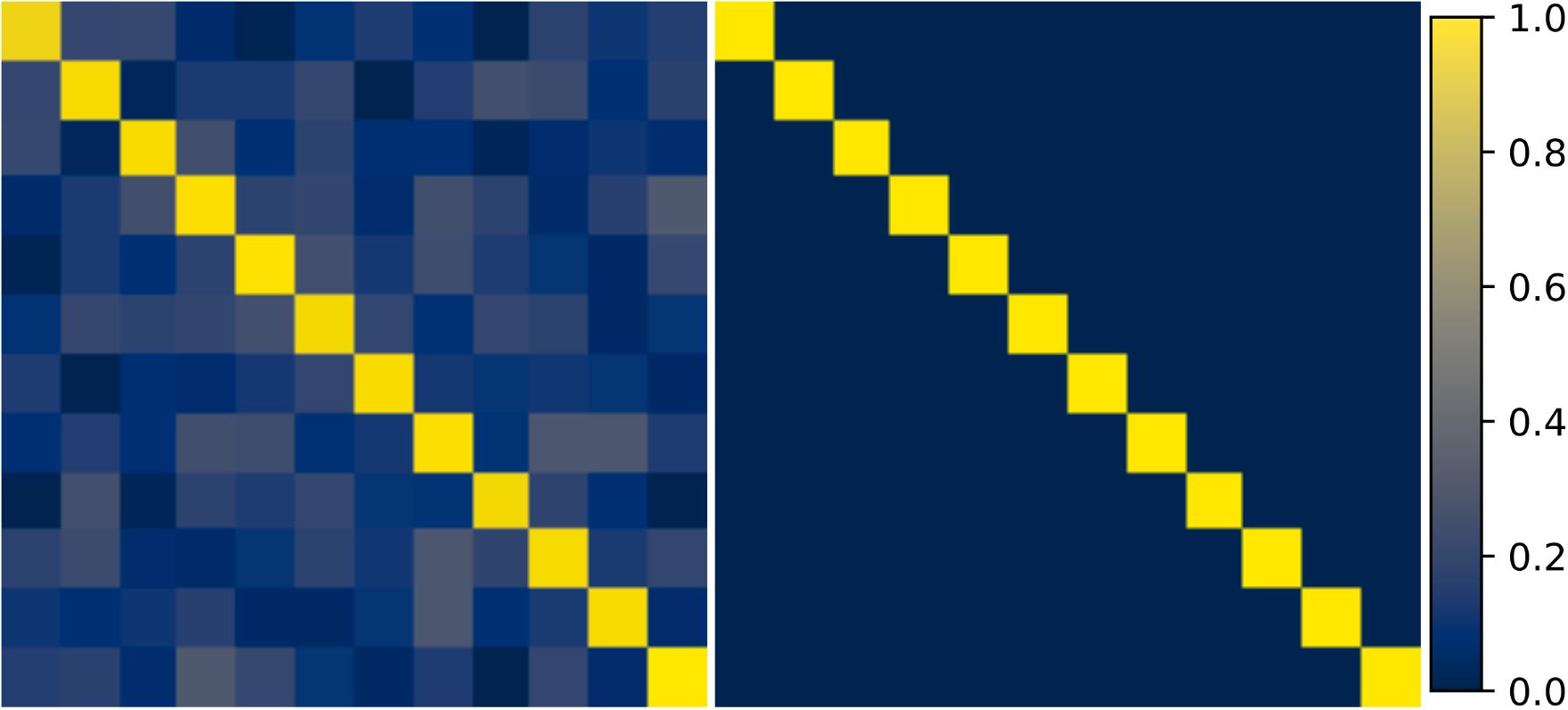}
        \caption{Illustration of the orthogonality of learnable proposal queries before (right) and after (left) training.}
        \label{query}
\end{figure}

\subsection{Ablation Study}
\label{abl}
Next we conduct various ablation experiments to understand the contribution of each individual component in our design. As shown in Table~\ref{tab:abl}, model (a) makes prediction using $\mathring{F}_{\text{hist}}$, and model (b) using $E_{\text{prop}}$. Since $E_{\text{prop}}$ are generated based on $F_{\text{hist}}$ that encodes the full historical states of a target agent, model (b) performs better than model (a). In model (c), the hierarchical encoding first applies $\mathring{F}_{\text{hist}}$ to attend to $\mathring{F}_{\text{rela}}$, and then attend to $\mathring{F}_{\text{road}}$. For model (d), the feature representation of a target agent in AcSR is directly used for prediction. Comparing models (c-d), we can see the advantage of a unified feature space for learning complex interactions over the modality-specific processing. By integrating $E_{\text{anch}}$ into model (d), model (e) improves the result, in particular for minFDE, suggesting the merit of learned anchors for endpoint prediction. In comparison to model (f) using a single prediction head, the full model (g) with hydra prediction heads gains further improvement.

\begin{table}[t]
  \centering
  \begin{tabular}{@{}lc@{}lc@{}lc@{}}
    \toprule
    Model & minADE$_1$ & & minFDE$_1$ \\
    \midrule
    (a) Compact History Encoding  & 3.31 & & 9.15 \\
    (b) Proposals  & 3.27 & & 9.06 \\
    (c) Hierarchical Encoding  & 2.04 & & 5.81\\
    (d) AcSR  & 2.02 & & 5.62 \\
    (e) AcSR with Anchors & 2.01 & & 5.47 \\
    (f) ProphNet-S & 1.98 & & 5.37 \\
    (g) ProphNet & 1.97 & & 5.33 \\
    \bottomrule
  \end{tabular}
  \caption{A variety of ablation studies and related design choices of ProphNet on the validation set of Argoverse-2.}
  \label{tab:abl}
\end{table}

\begin{figure}[t]
        \centering
        \includegraphics[width=\linewidth]{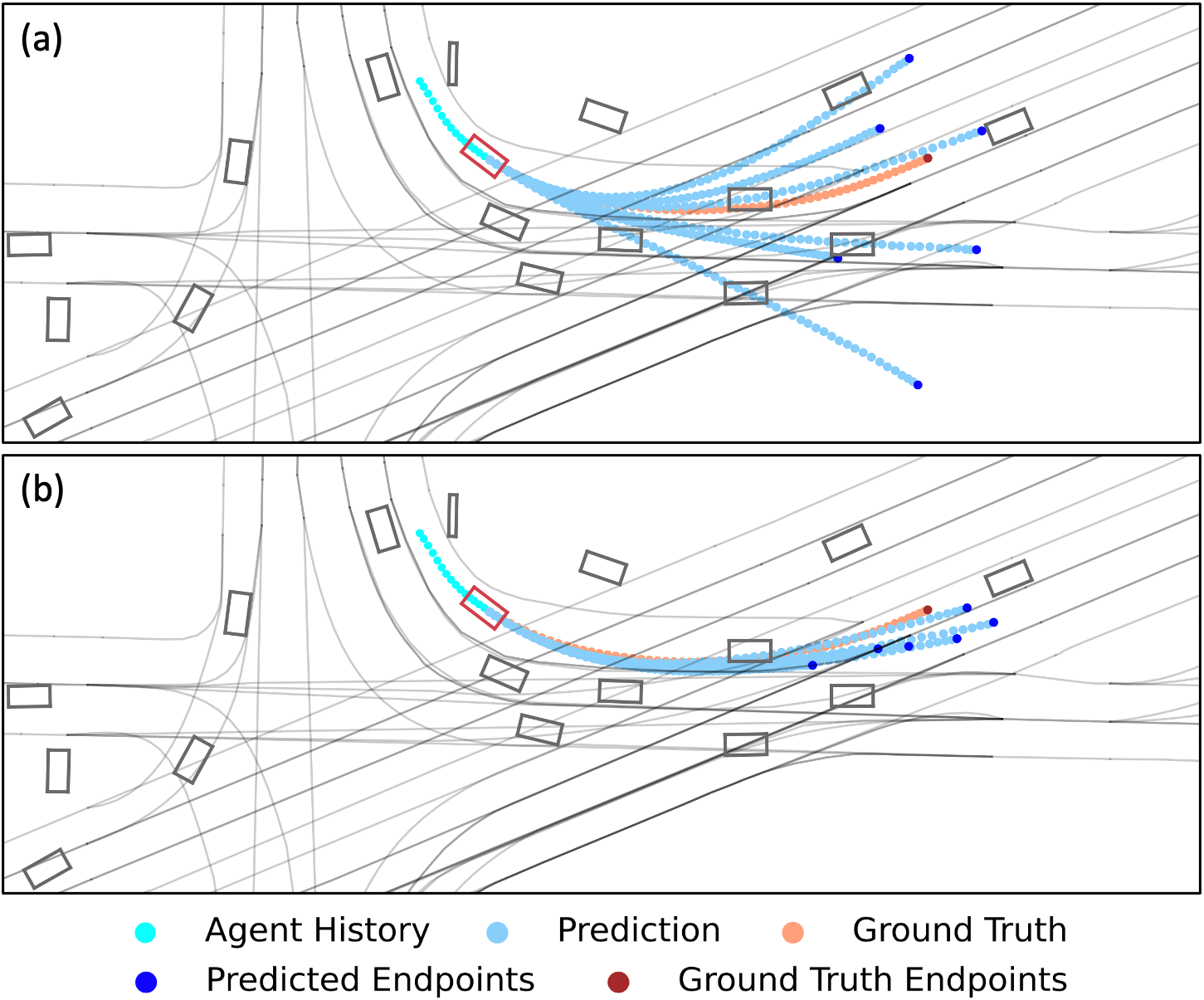}
        \caption{Illustration of the trajectories predicted (a) by the generated proposals directly and (b) by the anchor-informed proposals. }
        \label{prop}
\end{figure}

\subsection{Qualitative Results}

\noindent\textbf{Queries.} In Section~\ref{proposal-generation}, we introduce proposal queries that are orthogonally initialized for proposal generation to enhance output multimodality. As the queries are trainable, we investigate to what extent the learned queries can retain the orthogonality after training. As shown in Figure~\ref{query}, the initial queries are orthogonal, so their off-diagonal similarities are zero. After training, we find the similarities of the off-diagonal elements are still close to zero. This indicates that our model well maintains the diversity to induce proposals, showing no signs of mode collapse at this level.

\begin{figure*}[t]
        \centering
        \includegraphics[width=0.95\linewidth]{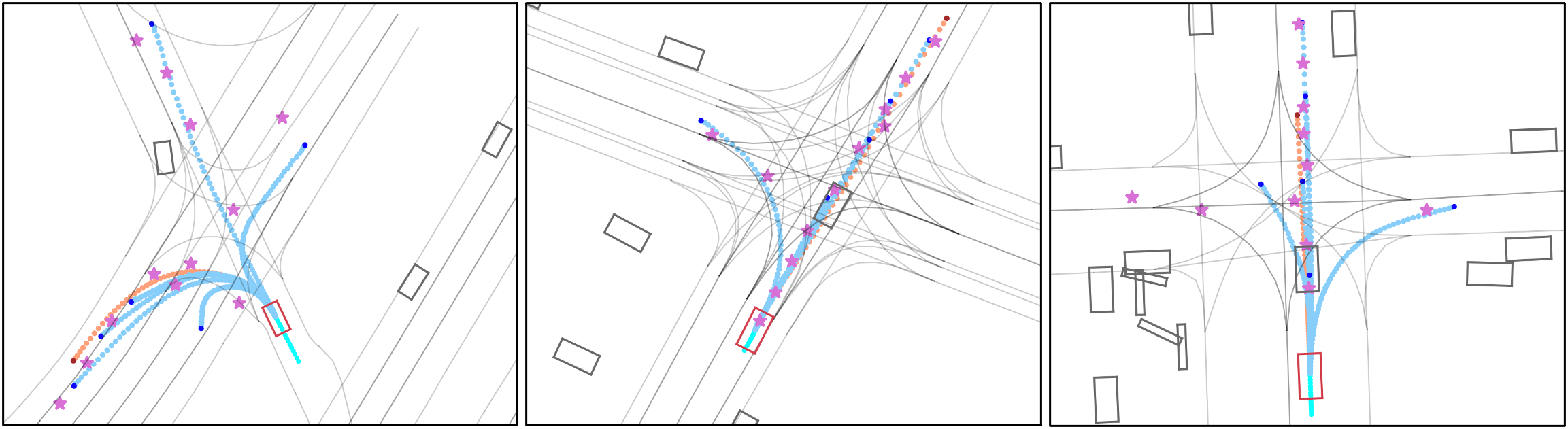}
        \caption{Illustration of the endpoints (pink stars) predicted by anchors in ProphNet. Note these endpoints are not used during inference, but are visualized here for better understanding of the learned anchors. The three examples also clearly demonstrate the diverse output multimodality enabled by our approach.}
        \label{anchor}
\end{figure*}

\begin{figure*}[t]
        \centering
        \includegraphics[width=0.95\linewidth]{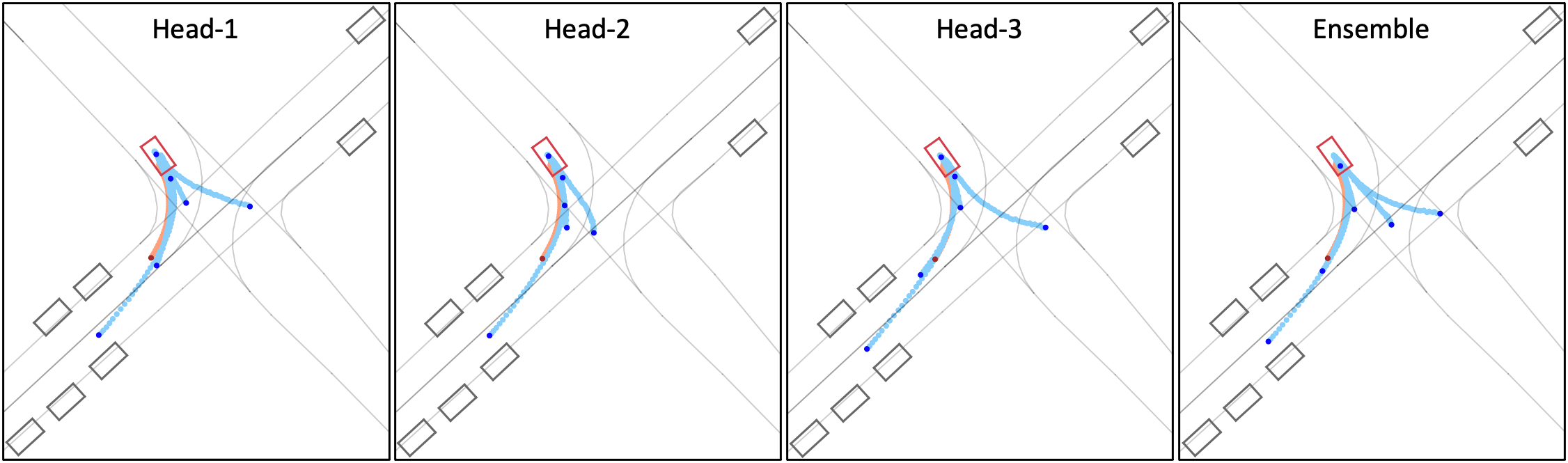}
        \caption{Illustration of the future trajectories output by hydra prediction heads and their ensemble in ProphNet.  }
        \label{pred}
\end{figure*}

\noindent\textbf{Proposals.} As illustrated in Figure~\ref{prop}(a), the trajectories predicted directly from proposals fit well to the agent history from the perspective of kinematics, though they are freely diverse or even off-road. This is because in our design the proposals are merely generated based on the agent dynamics, without the contextual information such as agent interactions and map constraints. It is found that the proposals generated in this way can be fairly diversified, catering to our purpose for rich coverage of potential motions at the early stage. When absorbing the goal-oriented scene context from anchors, the anchor-informed proposals produce accurate and multimodal future trajectories, as shown in Figure~\ref{prop}(b). The difference between (a-b) in this figure vividly presents the validity of the two components (i.e., proposals and anchors) in ProphNet.

\noindent\textbf{Anchors.} We make use of the supervision of endpoint estimation to guide the learning of anchors. Here we demonstrate the endpoints predicted from anchors in Figure~\ref{anchor}. Although such predicted endpoints are not used during inference of our approach, we visualize them for in-depth understanding of what is embedded in the learned anchors.
As illustrated in this figure, one can see that the predicted endpoints lie in road areas and mostly follow lane centerlines, which validates that the learned anchors have well encoded the scene context. Also, the predicted endpoints are positioned reasonably close to ground truth or the endpoints of other possible behavior modes, revealing the efficacy of anchor learning in our network.

\noindent\textbf{Hydra Prediction Heads.} Finally, we show each individual prediction result of the hydra heads in Figure~\ref{pred}. This is a challenging scenario as the target agent is static. We observe that the three prediction heads produce complementary diverse trajectories. After ensembling, the multimodal coverage of final output is further improved. These prediction results collectively verify our hydra-head design of randomly selecting subsets from AiP to enrich multimodality and encourage complementary prediction.

\section{Conclusion}
We have presented ProphNet, an agent-centric prediction model with anchor-informed proposals for efficient multimodal motion forecasting. 
ProphNet involves uniform encoding of heterogeneous input to construct a unified feature representation space AcSR, generating proposals and anchors to form AiP, and feeding AiP into hydra prediction heads to produce the multimodal output trajectories. 
Extensive experimental results on the two popular benchmarks demonstrate the effectiveness of our approach in terms of both prediction accuracy and inference latency.  
We hope this work would encourage more research toward practical network designs considered for real-world driving deployment, with not only high accuracy but also succinct architecture and efficient inference.

{\small
\bibliographystyle{ieee_fullname}
\bibliography{egbib}
}

\appendix
\section*{Appendix}

In Section~\ref{leaderboard}, we report additional comparisons with the leaderboard results on the two benchmarks. Section~\ref{attn} presents the visualization of learned attentions across multi-sourced input. Section~\ref{latency} describes the details regarding the latency evaluation of various methods. Section~\ref{viz} shows more qualitative results to illustrate the prediction of the proposed approach. Section~\ref{more_ablation} provides more ablation study.

\section{ProphNet on Leaderboards}
\label{leaderboard}

In the main paper, we have compared ProphNet with various published methods in Tables \ref{tab:Argo1ranking} and \ref{tab:Argo2}. Here we present more comparisons with the leading results (at the time of submission) including both published and unpublished methods that are reported on the leaderboards of Arogoverse-1\footnote{https://eval.ai/web/challenges/challenge-page/454/leaderboard/1279} and Argoverse-2\footnote{https://eval.ai/web/challenges/challenge-page/1719/leaderboard/4098}.  

To compare with the leaderboard results on Arogoverse-1, we follow the common practice of model ensembling and train six models with different random seeds. We then simply use non-maximum suppression to merge all predicted trajectories to produce the final prediction output. As demonstrated in Table~\ref{tab:Argo1}, ProphNet ranks the 1st among all 278 submissions. 

As for the leaderboard comparison on Arogoverse-2, we train three different models for ensembling, similar as that in Argoverse-1. As shown in Table~\ref{tab:Argo2r}, ProphNet ranks 2nd among all 32 submissions.

We then collect and compare the leading methods from the two leaderboards to see their generalizability across different benchmarks. As listed in Table~\ref{tab:Argo12}, one can find that only ProphNet and QCNet (unpublished) yield stable and superior performance on both leaderboards.  

To sum up, ProphNet achieves the 1st rank on the leaderboard of Argoverse-1 and the 2nd rank on the leaderboard of Argoverse-2. 
We have discussed the comparative study with published methods in the main paper. However, as the details of unpublished methods are not disclosed, we are unable to analyze the comparison with them qualitatively.

\begin{figure}[t]
    \centering
    \includegraphics[width=\linewidth]{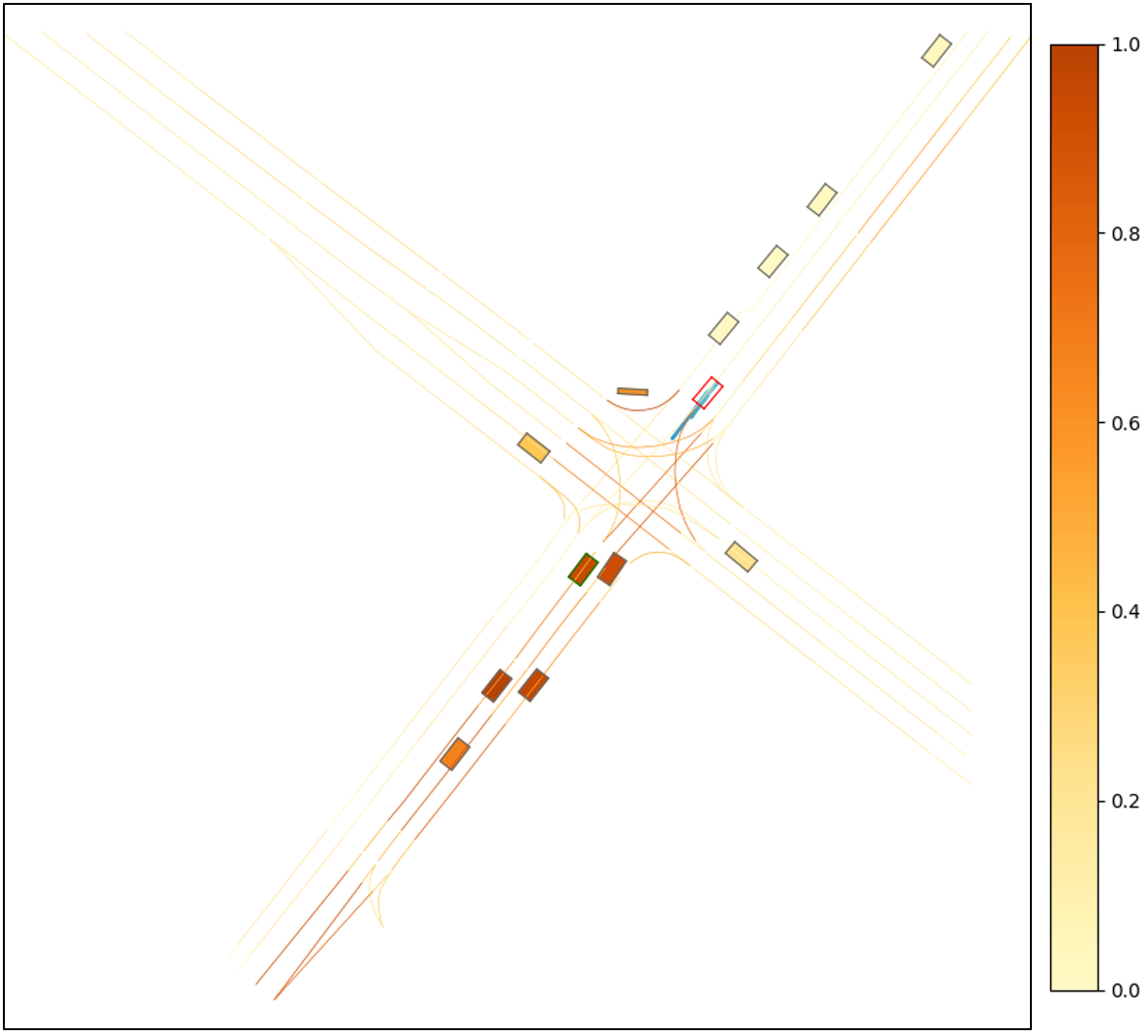}
    \vspace{-6mm}
    \caption{Illustration of the attention distribution of a target agent (red box) over the neighboring agents (ego-vehicle in green box) and road polylines learned by ProphNet.}
    \label{fig:attnviz}
\end{figure}

\section{Visualization of Learned Attentions}
\label{attn}
Next we visualize the learned attentions to multi-sourced input for in-depth understanding of the internal attention mechanism of ProphNet. Figure~\ref{fig:attnviz} illustrates the attention distribution of a target agent to other neighboring agents and road polylines. As shown in this figure, we observe that the target agent pays more attention to its frontal and lateral agents that would be potentially interacted with, while less attention to the agents behind. For road polylines, the target agent attends more on the frontal ones including the straight, left or right polylines that the future trajectories may lie on. This attention distribution learned in ProphNet is reasonably alike how humans attend to different traffic elements at an intersection.

\section{Details on Latency Evaluation}
\label{latency}
In this section, we describe the details of evaluating inference latency shown in Figure~\ref{intro} and Table~\ref{tab:latency}. As pointed out in the main paper, in addition to prediction accuracy, inference latency is an equally important measurement of a motion forecasting model for real-world driving deployment. Table~\ref{source} lists the sources that are used to conduct the inference latency evaluation. We employ the open-sourced code and models, and evaluate them under the same setting (tested with a single NVIDIA V100 GPU and the number of agents set to 64). As a special case, since there is no open-sourced material of Wayformer~\cite{wayformer}, we implement the model following the specifications detailed in the appendix (Table 5) of related paper~\cite{wayformer}.  

\begin{table*}[t]
  \centering
  \begin{tabular}{@{}lc@{}lc@{}lc@{}lc@{}lc@{}lc@{}lc@{}lc@{}lc@{}lc@{}lc@{}}
    \toprule
    Method &minADE$_6$& &minFDE$_6$& &minADE$_1$& &minFDE$_1$& &MR& &brier-minFDE\\
    \midrule
    Wayformer~\cite{wayformer} &0.77& &1.16& &1.64& &3.66& &0.12& &1.74 \\
    VI LaneIter*  &0.77 & &1.11& &1.52& &3.28& &\textbf{0.11}& &1.73 \\
    FFINet*  &0.76& &1.12& &1.53& &3.36& &\textbf{0.11}& &1.73\\
    QCNet*  &\textbf{0.74}& &\textbf{1.07}& &1.54& &3.37& &\textbf{0.11}& &1.70 \\
    ProphNet &0.76& &1.13& &\textbf{1.49}& &\textbf{3.26}& &\textbf{0.11}& &\textbf{1.69} \\
    \bottomrule
  \end{tabular}
  \vspace{-1mm}
  \caption{Comparison of the top 5 results of both published and unpublished methods on the leaderboard of Argoverse-1 by the date November 18, 2022. Note brier-minFDE is the primary ranking metric. * denotes unpublished methods. }
  
  \label{tab:Argo1}
\end{table*}

\begin{table*}[h]
  \centering
  \begin{tabular}{@{}lc@{}lc@{}lc@{}lc@{}lc@{}lc@{}lc@{}lc@{}lc@{}lc@{}lc@{}}
    \toprule
    Method &minADE$_6$& &minFDE$_6$& &minADE$_1$& &minFDE$_1$& &MR& &brier-minFDE\\
    \midrule
    OPPred* &0.71& &1.36& &1.79& &4.61& &0.19& &1.92\\
    TENET~\cite{tenet}  &0.70& &1.38& &1.84& &4.69& &0.19& &1.90 \\
    GNet*  &0.69& &1.34& &1.72& &4.40& &0.18& &1.90\\
    ProphNet &0.66& &1.31& &1.78& &4.80& &0.17& &1.89 \\
    QCNet* &\textbf{0.62}& &\textbf{1.19}& &\textbf{1.56}& &\textbf{3.96}& &\textbf{0.14}& &\textbf{1.78} \\
    \bottomrule
  \end{tabular}
  \vspace{-1mm}
  \caption{Comparison of the top 5 results of both published and unpublished methods on the leaderboard of Argoverse-2 by the date November 18, 2022. Note brier-minFDE is the primary ranking metric. * denotes unpublished methods. }
  \label{tab:Argo2r}
\end{table*}

\begin{table*}[!h]
  \centering
  \begin{tabular}{@{}lc@{}lc@{}lc@{}lc@{}lc@{}lc@{}lc@{}}
    \toprule
    Method &Rank (Argo-1)& &Rank (Argo-2)& &brier-minFDE (Argo-1)& & brier-minFDE (Argo-2)\\
    \midrule
    Proln* &6& &7& &1.75& &1.93\\
    TENET~\cite{tenet}  &11& &4& &1.77& &1.90 \\
    GNet*  &7& &3& &1.75& &1.90\\
    VI LaneIter* &4& &10& &1.73& &2.00\\
    QCNet* &2& &1& &1.70& &1.78&  \\
    ProphNet &1& &2& &1.69& &1.89 \\
    \bottomrule
  \end{tabular}
  \vspace{-1mm}
  \caption{Comparison of the top ranking results of both published and unpublished methods across the leaderboards of Argoverse-1 and Argoverse-2 by the date November 18, 2022. * denotes unpublished methods.}
  \label{tab:Argo12}
\end{table*}

\section{More Qualitative Results}
\label{viz}
\begin{figure*}
    \centering
    \includegraphics[width=\textwidth]{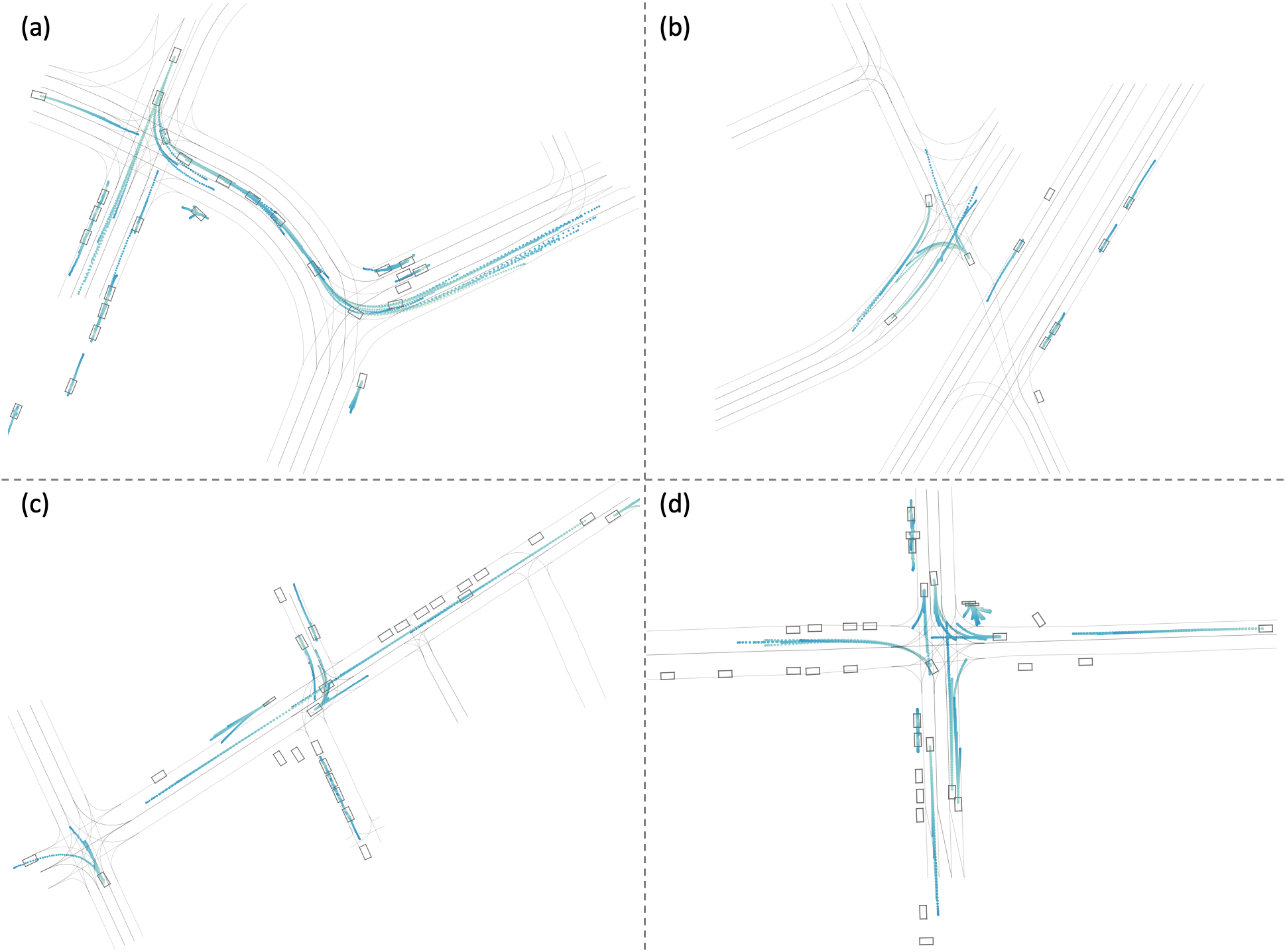}
    \caption{Illustration of the multimodal future trajectories predicted by ProphNet at various challenging scenarios with complex road topologies (a-b) and crowded intersections (c-d).}
    \label{fig:allviz}
\end{figure*}

We further provide more visualization of the predicted trajectories in challenging scenarios. Figure~\ref{fig:allviz} demonstrates four representative scenes with the holistic prediction output by ProphNet. As shown in this figure, (a-b) depict the complex road topologies, where ProphNet is able to produce multiple accurate trajectories that align to the lane centerlines reasonably well; and (c-d) illustrate the crowded intersections, where ProphNet predicts rich and rational multimodal future trajectories.

\section{More Ablation Study}
\label{more_ablation}
\begin{figure}[h]
        \centering
        \includegraphics[width=\linewidth]{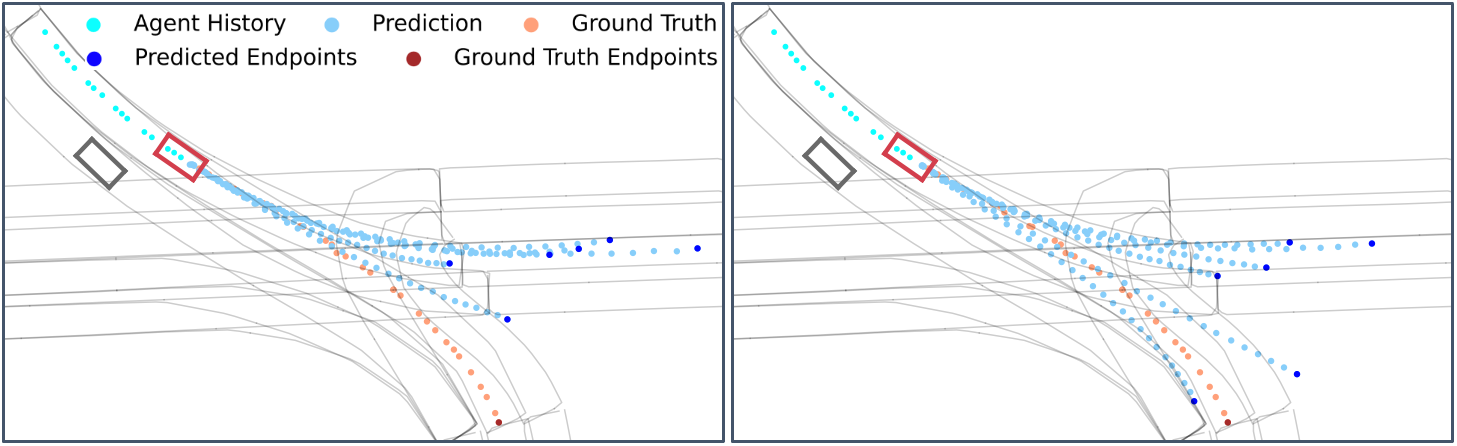}
        \caption{Illustration of the predicted trajectories by ProphNet-S with randomly (left) and orthogonally (right) initialized queries.}
        \label{rand_orth}
\end{figure}

Table~\ref{more-abl} provides more ablation studies for deeper understanding of our approach. We first compare different ways for initializing proposal queries in (a) and (f), where the random initialization is observed to be inferior to the orthogonal initialization. Figure~\ref{rand_orth} further shows the qualitative comparison. We then compare different ways of fusing proposals and anchors in (b) and (f). It is found that the simple summation performs better than the attention that incurs a higher computation cost. We next compare the direct use of anchor points with the proposed anchor embeddings in (c) and (f), and find that the latter is superior, validating the rich goal-oriented contexts encoded in the anchor embeddings. Finally, we compare different ways of input sequence encoding in (d-f), where gMLP outperforms both MLP and 1D-CNN.    
As for the failure cases of our approach, we find some long-horizon trajectories predicted in challenging scenes (e.g., roundabout) are occasionally off the road. One solution to mitigating this issue is to enforce an off-road penalty in training. 

\begin{table}[h]
  \centering
  \begin{tabular}{@{}lc@{}lc@{}lc@{}lc@{}}
    \toprule
    Model & minADE$_1$& &minFDE$_1$ \\ 
    \midrule
    (a) Random Initialization  &1.30& &2.81\\
    (b) Fusion by Attention  &1.29& &2.81\\
    (c) Use of Anchor Points  &1.29& &2.80\\
    (d) MLP Encoding  &1.40& &3.02\\
    (e) 1D-CNN Encoding  &1.31& &2.83\\
    (f) ProphNet-S  &\textbf{1.28}& &\textbf{2.77}\\
    \bottomrule
  \end{tabular}
  \vspace{-1mm}
  \caption{A variety of additional ablation studies on Argoverse-1.}
  \label{more-abl}
\end{table}

\begin{table*}[t]
  \centering
  \begin{tabular}{p{0.19\textwidth}p{0.8\textwidth}}
    \toprule
    Method  &Source\\
    \midrule
    VectorNet~\cite{vectornet} &\href{https://github.com/Henry1iu/TNT-Trajectory-Prediction}{https://github.com/Henry1iu/TNT-Trajectory-Prediction}\\
    LaneGCN~\cite{liang2020learning}  &\href{https://github.com/uber-research/LaneGCN}{https://github.com/uber-research/LaneGCN}\\
    mmTransformer~\cite{liu2021multimodal}  &\href{https://github.com/decisionforce/mmTransformer}{https://github.com/decisionforce/mmTransformer}\\
    DenseTNT~\cite{densetnt}  &\href{https://github.com/Tsinghua-MARS-Lab/DenseTNT}{https://github.com/Tsinghua-MARS-Lab/DenseTNT} \\
    MultiPath++~\cite{mp++} &\href{https://github.com/stepankonev/waymo-motion-prediction-challenge-2022-multipath-plus-plus.}{https://github.com/stepankonev/waymo-motion-prediction-challenge-2022-multipath-plus-plus} \\
    Wayformer~\cite{wayformer} &Our implementation following the model and training details in the appendix (Table 5) of~\cite{wayformer}\\
    \bottomrule
  \end{tabular}
  \caption{Sources of different methods used for inference latency evaluation.}
  \label{source}
\end{table*}

\end{document}